\documentclass[10pt,twocolumn,letterpaper]{article}

\usepackage{wacv}              % To produce the CAMERA-READY version
\usepackage{graphicx}
\usepackage{amsmath}
\usepackage{amssymb}
\usepackage{booktabs}
\usepackage{svg}
\usepackage{multirow}

\usepackage[pagebackref,breaklinks,colorlinks]{hyperref}

\usepackage[capitalize]{cleveref}
\crefname{section}{Sec.}{Secs.}
\Crefname{section}{Section}{Sections}
\Crefname{table}{Table}{Tables}
\crefname{table}{Tab.}{Tabs.}

\def\wacvPaperID{00019} % *** Enter the WACV Paper ID here
\def\confName{WACV}
\def\confYear{2025}

\begin{document}

%%%%%%%%% TITLE - PLEASE UPDATE
\title{Uncertainty and Energy based Loss Guided Semi-Supervised Semantic Segmentation}

\author{Rini Smita Thakur, Vinod K Kurmi\\
Indian Institute of Science Education and Research Bhopal, India\\
%Institution1 address\\
{\tt\small \{rinithakur, vinodkk\}@iiserb.ac.in}
% For a paper whose authors are all at the same institution,
% omit the following lines up until the closing ``}''.
% Additional authors and addresses can be added with ``\and'',
% just like the second author.
% To save space, use either the email address or home page, not both
% \and
% Vinod K Kurmi\\
% IISER, Bhopal\\
% %First line of institution2 address\\
% {\tt\small vinodkk@iiserb.ac.in}
% 
}
\maketitle

%%%%%%%%% ABSTRACT
\begin{abstract}
   Semi-supervised (SS) semantic segmentation exploits both labeled and unlabeled images to overcome tedious and costly pixel-level annotation problems. Pseudolabel supervision is one of the core approaches of training networks with both pseudo labels and ground-truth labels. This work uses aleatoric or data uncertainty and energy based modeling in intersection-union pseudo supervised network.The aleatoric uncertainty is modeling the inherent noise variations of the data in a network with two predictive branches. The per-pixel variance parameter obtained from the network gives a quantitative idea about the data uncertainty. Moreover, energy-based loss realizes the potential of generative modeling on the downstream SS segmentation task. The aleatoric and energy loss are applied in conjunction with pseudo-intersection labels, pseudo-union labels, and ground-truth on the respective network branch. The comparative analysis with state-of-the-art methods has shown improvement in performance metrics. The code is availaible at \href{https://visdomlab.github.io/DUEB/}{https://visdomlab.github.io/DUEB/}.
\end{abstract}

%%%%%%%%% BODY TEXT
\section{Introduction}
\label{sec:intro}

Semantic segmentation (SS) is an important branch of computer vision, which finds application in medical imaging, autonomous driving, and other intelligent industrial systems. There is significant performance improvement with the advent of supervised deep learning methodologies with extensive labeled data~\cite{segreview}. However, in many practical scenarios labeled data scarcity is the major problem. The constraint for supervised image segmentation is the requirement of enormous pixel-label annotated data, whose acquisition is tedious and time-consuming. It takes around 1.5 hours to annotate a single high-resolution urban street scene (Cityscapes dataset)~\cite{paper1}, and sixty minutes for camouflaged object image~\cite{paper2} (COD10K dataset). So, there is a gradual shift from fully supervised approaches to weakly supervised and SS approaches. The weakly supervised methods leverage image-level labels with bounding boxes or scribbles, whereas SS methods leverage both labeled and unlabeled images. The SS semantic methods are categorized into adversarial methods, consistency regularization, contrastive learning, pseudo-labeling, and hybrid methods~\cite{2023survey,2020survey}. Pseudo-labeling is one of the most useful, simple and effective methods in SS segmentation.  Pseudo-labeling implies the generation of pseudolabels from the supervised trained network and re-training with the new dataset composed of ground truth and pseudolabels. 
% To insert a figure: \input{figs/template}
% Or table: \input{tables/template}
 \begin{figure}[t]
\includegraphics[width=0.50\textwidth]{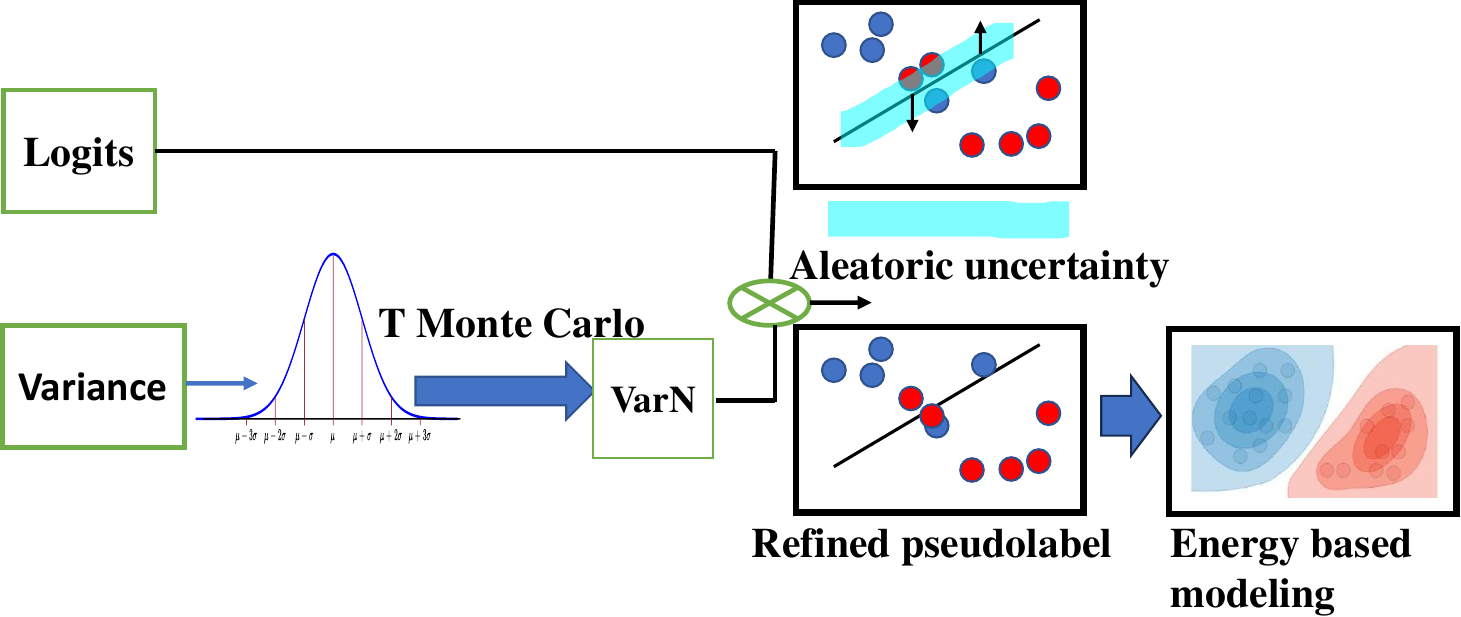}
\caption{Refinement of pseudolabel with aleatoric uncertainty and energy based modeling}
\label{fig:main}
\end{figure}

  There is always a quest for the refinement of pseudo labels as the network is prone to confirmation bias. Uncertainty quantification is the preferred approach for refining pseudolabels, relying on two fundamental types of uncertainty: epistemic (model) uncertainty, which reflects the variability in model parameters, and aleatoric (data) uncertainty, which captures the inherent noise present in the input data. There are various test-time epistemic uncertainty (model) estimation techniques for pseudolabel refinement~\cite{sfda}. However, the inclusion of aleatoric (data) uncertainty in the Bayesian framework during training has not been explored in the SS segmentation setup. The non-inclusion of training time aleatoric uncertainty can be attributed to the fact that it requires significant architectural change for per-pixel variance calculation, which is not always feasible~\cite{aleamodi}. Secondly, appropriate change in the training process is required for variance optimization according to the specific computer vision problem.  On the other hand, epistemic uncertainty can be modeled at the inference time even in pre-trained models with a more generalized approach without cost modification. Therefore, there is a huge surge in epistemic methods such as thresholding, entropy minimization, ensembling and  Monte-Carlo dropout with Bayesian approximation~\cite{different}, but aleatoric inclusion is a neglected domain. So, we hypothesize to include the effect of object boundaries, inter and intra-class variation, and low-contrast regions with training-time aleatoric uncertainty quantification.

  % The inclusion of uncertainty measure for pseudolabeling is done by softmax-based confidence, entropy, Monte-Carlo dropout using Bayesian approximation, ensemble-based estimation and data augmentation based methods ~\cite{different}. The existing methods such as thresholding, entropy minimization, and Monte-Carlo dropout cater to model uncertainty at the inference time. On the other hand, other non-inference time methods such as data augmentation strategies and ensembling are computationally expensive. There are various epistemic uncertainty (model) estimation techniques for pseudolabel refinement at the inference time ~\cite{sfda}. However, the inclusion of aleatoric (data) uncertainty in the Bayesian framework during training has not been explored in the SS segmentation setup. The data uncertainty captures the inherent noise observations in the input data itself. Our goal is to evaluate data uncertainty by employing different costs using the per-pixel variance information during training phase. 
  
  The other constraint with the SS segmentation methods  is the discriminative and deterministic framework, which fails to capture the generative distribution. The consistency regularized pseudo-supervised methods ~\cite{pseudoseg,crosspseudo,gct,mixedlabel,fan2023} leverage supervised cross-entropy loss with ground-truths, unsupervised cross-entropy loss with the pseudolabels and consistency loss to capture perturbations within discriminative and deterministic framework. However, the discriminative networks can be reinterpreted as energy based model (EBM) to include the advantages of generative networks, i.e., implicit modeling of input data $p(x)$ to avoid over-fitting with limited labeled training data, strong generalizability for out-of-detection sample, adversarial robustness, and improved uncertainty calibrations~\cite{advan,zhao2020joint}.

 In this study, we employ an EBM with a joint distribution of input images and labels of data to obtain generative functionality using a simple pseudo-labeled classifier. The discriminative classifier can be re-interpreted as the energy-based model for the joint distribution of input and ground truth~\cite{jem,Kurmi_2021_WACV}. The EBMs parameterize any multi-dimensional input into the scalar value.  

The joint energy based modeling and inclusion of certain uncertainty measures such as variance, entropy, dropout, etc., either at training or test time, can improve classification performance~\cite{dermouncertainty,alarab2023uncertainty,kendall}. Therefore, it has motivated us to design variance-based loss for the segmentation network in order to capture inherent noise variations of the input. Furthermore, additional energy-based loss can help the model in learning the joint distribution of data $p(x,y)$ leading to a more enhanced refinement. It favours aleatoric uncertainty quantification by implicitly modeling $p(x)$.

The SS segmentation network provides per-pixel variance for heteroscedastic uncertainty calculation. A normal distribution is placed over the variance, and this variance is added to the model's prediction. The heteroscedastic aleatoric loss incorporates this variance parameter along with the energy-based loss to further refine the results. Figure~\ref{fig:main} depicts the pseudolabel refinement with aleatoric uncertainty using both logits and variance. \\ 
The proposed framework, data uncertainty and energy loss (DUEB) is the pseudolabeling based SS segmentation network built upon the Conservative Progressive Collaborative Learning (CPCL) ~\cite{fan2023} framework. CPCL is a hybrid network that leverages consistency regularization and pseudo-labeling in a discriminative framework with two parallel branches. It generates union-intersection pseudo labels based on agreement and disagreement indicators on the outputs of two branches. These union and intersection help to generate better pseudo-labels. The main contributions of the DUEB are:
 % We propose a data uncertainty and energy-based loss (DUEB) driven SS semantic network that utilizes the power of generative models for better uncertainty calibration. In this framework, we utilize data uncertainty estimation during training time on two branches, termed as conservative and progressive catering to intersection and union pseudolabels respectively. Specifically, this work introduces aleatoric uncertainty-based loss attenuation and energy-based loss, along with pseudo-label refinement in the consistency regularized network. 
\begin{itemize}
    \item We propose the aleatoric data uncertainty-based loss framework with DeepLabv3+ ~\cite{chen2017rethinking}. We incorporate the distorted logit loss based on data-dependent variance and optimize the difference between undistorted logit and distorted logit loss.
\item Enhance the SS prediction by energy-based loss to incorporate generative modeling using the discriminative function.
\item Usage of union-intersection pseudo-labels for data uncertainty and energy loss calculation in an unsupervised setting and usage of ground truth in a supervised setting.
\end{itemize}
%-------------------------------------------------------------------------
\section{Related Work}
\label{sec:related}
\subsection{Semi-supervised segmentation}
The combination of consistency regularized network along with pseudo-supervision is widely used for the SS segmentation~\cite{pseudoseg,fixmatch,crosspseudo,mixedlabel,unpseudo}. The general methodology involves multiple identical networks or multiple networks with the same structure but different initialization with two or more predictive branches. Perturbations (input/feature/network) are applied, and corresponding pseudolabels are generated for all the branches. Cutout, CutMix, ClassMix and ComplexMix are commonly used image augmentation methods for input perturbations~\cite{cutmix}. \\Fixmatch~\cite{fixmatch}, Pseudoseg~\cite{pseudoseg}, and Unimatch ~\cite{yang2023revisiting} impose consistency on strongly and weakly augmented images. Fixmatch generates pseudolabels with thresholding, whereas PseudoSeg utilizes a calibrated fusion strategy. In PseudoSeg~\cite{pseudoseg}, the first network outputs decoder prediction and self-attention GradCam with weakly augmented image input. Thereafter, pseudolabels generated with calibrated fusion strategy supervise another network with strongly augmented image input. Unimatch~\cite{yang2023revisiting} is built upon Fixmatch, and it uses additional feature perturbations and two strong views perturbations, unlike Fixmatch. Cross Pseudo Supervision network (CPS)~\cite{crosspseudo} and S3MPL~\cite{mixedlabel} network forces consistency between two differently initialized networks ($f_{\theta1}$,  $f_{\theta2}$) with same augmented input. In CPS, the pseudo-hot label map of the first network $f_{\theta1}$ is used to supervise another network $f_{\theta2}$ and vice-versa. However, in S3MPL mixed pseudolabel map based on higher prediction confidence is used to supervise both  ($f_{\theta1}$,  $f_{\theta2}$). The SS medical segmentation uses uncertainty aware pseudolabels in a consistency regularized student-teacher framework~\cite{unpseudo}. The noisy pseudolabels are refined with uncertainty estimation by Kullback–Leibler variance. $U^2PL$~\cite{reliable} network proposes a methodology to sort reliable and unreliable pseudolabels based on entropy.\\
Recently, SS segmentation has progressed towards transformer-based architectures (e.g. Semi-CVT~\cite{huang2023semicvt}, AllSpark~\cite{wang2024allspark}). The extension of CPS is done in transformer-based architectures using ViT backbone in~\cite{li2023semi,li2023diverse}.

\subsection{Uncertainty Estimation}
Uncertainty quantification procedures are pivotal in uncertainty reduction during optimization and decision-making with two major approaches: Bayesian and Ensemble learning~\cite{abdar2021review}. The uncertainties in computer vision are classified as aleatoric (data) uncertainty and epistemic (model). Aleatoric uncertainty captures the inherent noise distributions in the input image pixels, whereas epistemic uncertainty captures the network uncertainty. The epistemic uncertainty is incorporated in SS segmentation works, whereas irreducible aleatoric uncertainty is a neglected domain. It is captured by putting prior distribution over the weights followed by variational Bayesian approximation over the weights posterior distribution. The epistemic uncertainty is incorporated in the segmentation network with Monte-Carlo dropout. The Monte Carlo dropout involves multiple forward passes turning off certain nodes as per dropout probability. The 3D SS left atrium student-teacher framework network exploits epistemic uncertainty information with predictive entropy metric to evaluate uncertainty aware consistency loss~\cite{epi1}. The other medical segmentation network identifies certain and uncertain areas based on the conservative-radical module with multiple decoders~\cite{epi2}. Kullback–Leibler variance is used for uncertainty estimation to rectify the noisy pseudolabels for SS medical image segmentation~\cite{unpseudo}. There are several methods to evaluate epistemic uncertainty with a single pass to reduce the computational burden with Monte-Carlo passes~\cite{mukhoti}.

Aleatoric uncertainty captures image noise, occlusions, blur, uncertain visual cues, etc. Heteroscedastic aleatoric uncertainty is modeled by Gaussian output models or Dirichlet models~\cite{al1}. The Gaussian output network models the noise at the logits into the Gaussian distribution. The mean of Gaussian distribution is equal to the output of the corresponding class, whereas variance is learned from the input image. The associated likelihood is obtained from the expectation of this probability by applying Monte Carlo to these distributions, drawing a certain number of samples from the Gaussian modeling of the logits. The resulting likelihood is formulated into the cross entropy loss to embed the aletoric uncertainty component. Aleatoric uncertainty is also applied during test-time augmentation, where the distribution of predictions is calculated with Monte-Carlo simulations with the prior distribution of image transformations and noise ~\cite{al2,kumar2021mitigating}.

\section{Method}
\label{sec:method}
In this section, we first define the problem statement, followed by a description of network architecture, data uncertainty and energy based loss in a union-intersection pseudolabel supervised framework. We have incorporated the variance layer in DeepLabv3+~\cite{chen2017rethinking} architecture and formulated aleatoric uncertainty and energy based loss in union-intersection pseudolabels supervised network.
\begin{figure*}[t]
        \centering
        \includegraphics[width=0.85\textwidth]{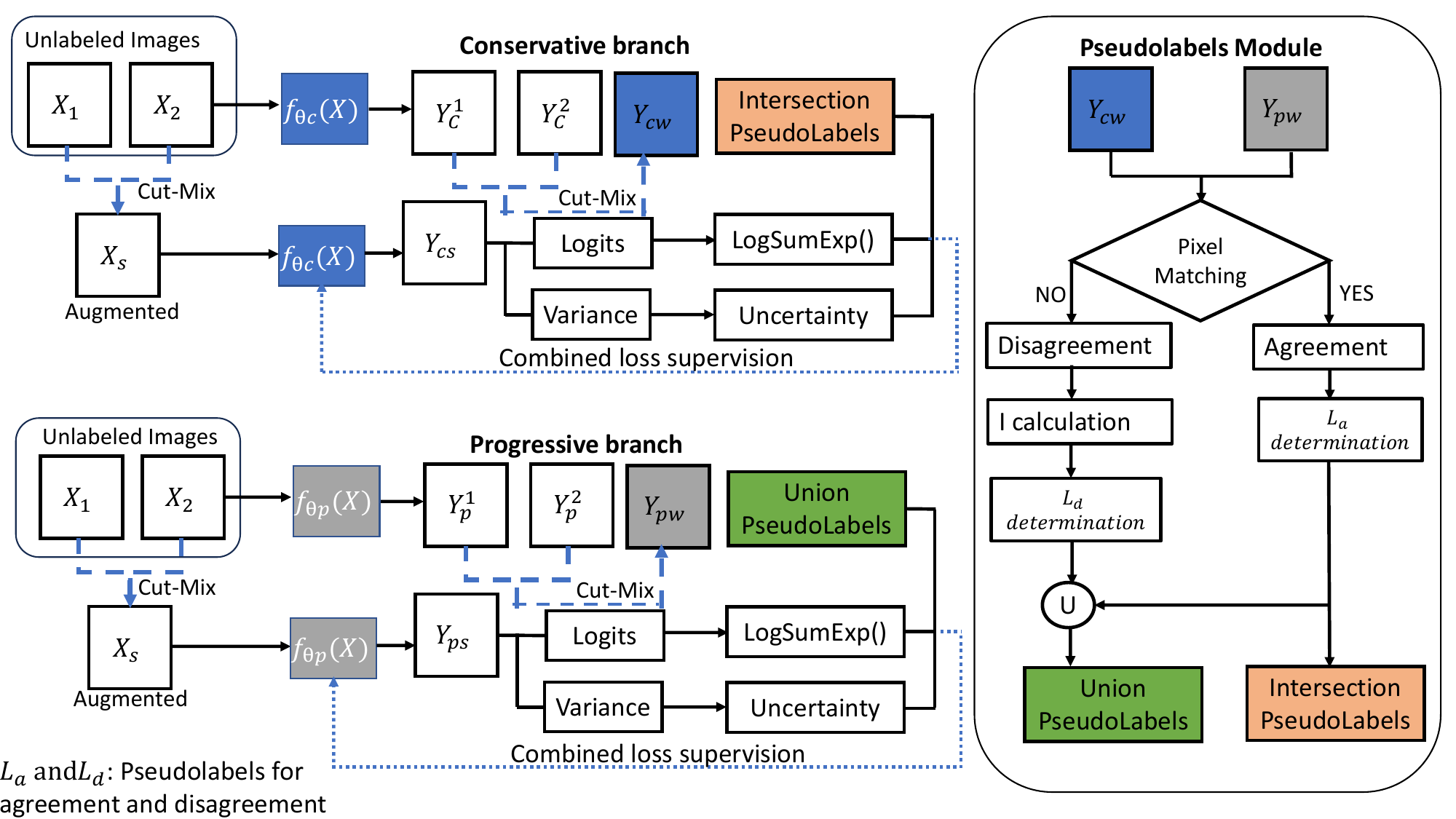}
        \caption{Block Diagram of proposed uncertainty and energy loss based framework with pseudolabels module for semi-supervised segmentation. Two branches: conservative and progressive are trained in parallel with intersection/union pseudolabels, data uncertainty loss and energy based loss.}
        \label{fig:main2}
    \end{figure*}
  \subsection{Problem Statement}
In SS semantic segmentation tasks, both labeled and unlabeled data are available, with the condition that the amount of unlabeled data is significantly greater than the labeled data. The labeled data with $P$ images and corresponding ground-truth are given by $D_{l}=\left\{(X_{l}^{1}, G_l^{1}),.......,(X_{l}^{P}, G_l^{P})\right\}$. The $Q$ unlabeled images is given by $D_{u}=\left\{X_{u}^{1}, .......,X_{u}^{Q}\right\}$.  A large amount of unlabeled data yields good segmentation results on test data. In this work,  we follow similar architecture details presented in~\cite{fan2023}. The main framework consists of two segmentation networks, $f{( ;\theta}_{c})$ and $f{( ;\theta}_{p})$ with the same structure but different initialization. It utilizes labeled data in a traditional supervised manner and unlabeled data with union-intersection pseudo labels. Both networks predict the pixel-wise semantics segmentation class labels and pixel-wise variance (uncertainty) values.  
\subsection{Network Architecture}
We propose a novel SS segmentation framework that incorporates aleatoric or data uncertainty-based loss and energy-based loss as shown in Figure~\ref{fig:main2}. The framework consists of two networks $f_{\theta_c}{( ;\theta}_{c})$ and $f_{\theta_p}{( ;\theta}_{p})$, termed as the conservative branch and progressive branch, respectively. $\theta_{c}$ and $\theta_{p}$ are the parameters of these networks respectively. Each network produces logits and variance at the output. The supervised learning is done with traditional categorical cross-entropy in which the same labeled images are given as the input in both networks. In an unsupervised setting, two unlabeled images $X_{1}$ and $X_{2}$ are given as input to both $f_{\theta_c}{( ;\theta}_{c})$ and $f_{\theta_p}{( ;\theta}_{p})$.The CutMix method is adopted for generating the strongly augmented image from individual images $a$ and $b$ to generate the output pixel based on the random mask $m$ given in Eq.~\ref{eq:1}:
\begin{equation}
mix(a,b,m)=(1-m)\odot a + m \odot b \label{eq:1}
\end{equation}
The strongly augmented image $X_{s}$ is obtained from Eq.~\ref{eq:2}.
\begin{equation}
X_{s}=mix(X_{1},X_{2},m) \label{eq:2}
\end{equation}
The images $X_{1}$, $X_{2}$ and $X_{s}$ are given as inputs in the conservative branch to generate $Y_{c}^1$, $Y_{c}^2$ and $Y_{cs}$ given by Eq.~\ref{eq:3}, \ref{eq:4} and \ref{eq:5}
\begin{equation}
Y_c^1\leftarrow \  \arg \max_{y}  f_{\theta_{c}}(y|X_{1})
\label{eq:3}
\end{equation}
\begin{equation}
Y_c^2\leftarrow \  \arg \max_{y}  f_{\theta_{c}}(y|X_{2})
\label{eq:4}
\end{equation}
\begin{equation}
Y_{cs}\leftarrow \  \arg \max_{y}  f_{\theta_{c}}(y|X_{s})
\label{eq:5}
\end{equation}
The output $Y_{c}^1$, $Y_{c}^2$ are combined at the output using the same CutMix procedure as that of input to give $Y_{cw}$ using the same mask $m$, i.e.  $mix$($Y_{c}^1$, $Y_{c}^2$,$m$). 
Similarly, we can get the output of the progressive branch as $Y_{ps}$ and $Y_{pw}$ ($mix$($Y_{p}^1$, $Y_{p}^2$,$m$)) with the network $f{(\theta}_{p})$.

The mixed output of both networks $(Y_{cw}, Y_{pw})$ obtained by mixing two individual images output is used for the generation of pseudo labels. These pseudo labels supervise the branches whose input is the Cut-Mix version of two images. The pseudo labels of the conservative branch are obtained by the intersection between individual pixels of $Y_{cw}$ and $Y_{pw}$. Whereas the pseudo labels for the progressive branch are formed by the union operation of $Y_{cw}$ and $Y_{pw}$. The union operation combines pixels using both agreement and disagreement indicators. \\
The intersection pseudo labels $y_{\text {inter }}^i, $ are obtained by per-pixel comparison of $Y_{cw}$ and $Y_{pw}$, as both outputs should be consistent with each other. The union pseudolabels $y_{\text {union}}^i,$ consider both agreement and disagreement of pixels in $Y_{cw}$ and $Y_{pw}$.\\
 Usually, the disagreement of the pixels denotes that the pixel is uncertain as both networks give different predictions.There is always a trade-off between selecting high-quality pseudolabels and utilizing all the pseudolabels. The prediction-confidence based approach used in some of the recent works avoids performance degradation at the expense of wastage of unlabeled data. In~\cite{hung}, only $27-36 \%$ pixels of the unlabeled Cityscapes dataset are used. Therefore, in the case of disagreement, we chose the class with a higher disagreement indicator as it is a more difficult class to predict.  The class-wise disagreement indicator (I) given in CPCL~\cite{fan2023} evaluates the pseudo labels of the disagreement regions. The final union pseudo labels are the union of the agreement and disagreement pixels between two predictions, thereby leveraging a high quantity of pseudolabels. The intersection pseudolabels supervise $f_{\theta_c}{( ;\theta}_{c})$, whereas union pseudolabels supervise $f_{\theta_p}{( ;\theta}_{p})$ with Cut-Mix strongly augmented image as the input $X_{s}$. A detailed explanation of class-wise disagreement indicator and dynamic confidence-based weight $w_u^i$ is given in Supplementary.
The loss function with intersection and union pseudolabels is given as:

    \begin{equation}
     \mathcal{L}_{int}=\frac{1}{N} \sum_i^{N}  w_u^i \operatorname{CE}\left( f_{\theta_c}^i\left(X_s\right);y_{\text {inter }}^i\right) 
\end{equation}

    \begin{equation}
     \mathcal{L}_{uni}=\frac{1}{N} \sum_i^{N}  w_u^i \operatorname{CE}\left( f_{\theta_p}^i\left(X_s\right);y_{\text {union }}^i\right) 
\end{equation}
$f_{\theta_c}^i(X_s)$ and $f_{\theta_p}^i(X_s)$ are the $i^{th}$ pixel output of given input image $X$. 
For the supervised data:   \begin{equation}
     \mathcal{L}_{sup}=\frac{1}{N} \sum_i^{N}  \operatorname{CE}\left( f_{\theta}^i\left(X_l\right);G_l^i\right) 
\end{equation}
These losses can be combined as:
\begin{equation}
    \mathcal{L}_{det}=\mathcal{L}_{sup}+\gamma_{int}\mathcal{L}_{int}+\gamma_{uni}\mathcal{L}_{uni}
\end{equation}

$\gamma_{int}$ and $ \gamma_{uni}$ are the hyperparameter for union and intersection supervision based loss function. These pseudo labels obtained using the union and interaction from two networks do not consider the uncertainty associated with the data. Thus we propose the following loss function to further enhance the pseudo-label supervision. 
\subsection{Data Uncertainty Estimation}
Both the networks share the same segmentation structure as DeepLabv3+~\cite{chen2017rethinking} with different weights initialization. The network's last layer is modified to give a variance layer in the output of the network along with the predicted logits. The variance layer is the convolutional layer which is applied over the features of the last layer. Thus, the model outputs per-pixel variance along with the logits. The Bayesian aleatoric uncertainty loss function trains the variance layer's parameter. We follow similar to ~\cite{kendall,joshi2021data} to obtain the variance and data uncertainty in the deep model.\\
For any given input $X$, corresponding label $Y$ and model $f(;\theta)$, the data uncertainty can be obtained from following:
\begin{equation}
\hat{Y},\sigma_x= f (X;\theta)
\label{eq:11}
\end{equation}
\begin{equation}
\operatorname{\mathit{diff}} =\mathit{CE}(\mathit { \hat{Y}, Y })-\mathit{CE}\left(\operatorname{\mathit{\hat{Y}}}+ \epsilon_{\mathrm{t}}{, Y }\right) \text {; }   \epsilon_{\mathrm{t}} \sim \mathcal{N}(0, \sigma_x)
\label{diff}
\end{equation}
Here $\sigma_x$ is the pixels-wise variance for input $X$ and $\hat{Y}$ is the logit output for given input $X$ and model$f(;\theta)$.
$CE(,)$is the cross-entropy loss function.  $\mathcal{N}(0, \sigma_x)$ is the normal distribution with mean $0$ and $\sigma_x$ is data dependent variance.\\
In the aleatoric loss, variance-based components are added to the original undistorted cross entropy loss $l_u$ which is obtained from the model without variance layer. Apart from $l_u$, the other component reduces the difference between cross-entropy loss on the original logits and cross-entropy loss on the distorted logits. The distorted logits have Gaussian noise sampled according to the predicted variance. Gaussian noise is sampled from the normal distribution with mean zero and standard deviation equivalent to the square root of the predicted variance. Monte Carlo (MC) sampling generates multiple distorted versions of the logits by adding Gaussian noise. The difference tends to reduce the data uncertainty as it tends to optimize the variance with respect to undistorted logits. This difference
is passed through the exponential linear unit for negative
value handling. The expression $e^{\sigma^2}$
 exponentially scales the variance. The exponential function is chosen because it amplifies larger values of variance more significantly while maintaining smooth behaviour for smaller values. This means the model will penalize larger uncertainties more heavily, encouraging it to reduce uncertainty wherever possible.
 % A normal distribution with mean zero and standard deviation equivalent to the square root of the predicted variance is created to obtain the aleatoric uncertainty loss function. Using the Monte Carlo (MC) sampling methods, the predicted logits are distorted, which is equivalent to adding noise parameters as per the variance of the data. Therefore, the distorted loss is the application of categorical cross-entropy on the logits and samples Gaussian noise (its variance is data dependent). We tend to optimize the difference between undistorted loss and distorted loss. This difference is passed through the exponential linear unit for negative value handling. 
 The overall data uncertainty loss ${l}_{d u}$  is given by Eq.~\ref{luncer}:
\begin{equation}
\operatorname{\mathcal{L}_{ale}}=\frac{1}{T} \sum_T\left[(-E L U * \operatorname{\mathit{diff}}) * l_u+l_u+\left(e^{\sigma^2}-1\right)\right]
\label{luncer}
\end{equation}

where $l_u$ is the undistorted loss, i.e., cross-entropy loss ($CE$) between ground-truth ($Y$) and $\hat{Y}$ i.e. $l_u=\mathit{CE}(\mathit { \hat{Y}, Y })$, $T$ are Monte-Carlo samples, $ELU$ is an exponential linear unit.
 For the conservative and progressive branches, we can define the uncertainty loss as $\mathcal{L}^c_{ale}$ and $\mathcal{L}^p_{ale}$, respectively. Note that for calculating the  $\mathcal{L}^c_{ale}$, we use $y_{inter}$ as the pixel label. Similarly, $y_{union}$ are used as the pixel label for obtaining  $\mathcal{L}^p_{ale}$.
 \begin{table*}[]
\footnotesize
\centering
\scalebox{1}{
\begin{tabular}{|c|c|c|c|c|c|c|c|}
\hline
\textbf{Partition} & \textbf{Methods} & \textbf{mIoU}  & \textbf{Animal} & \textbf{Vehicle} & \textbf{Indoor} & \textbf{Person} & \textbf{Background} \\ \hline
\multirow{3}{*}{1/2}                & Supervised                        & 74.05                           & 84.45                            & 76.90                              & 55.76                            & 81.95                            & 93.57                                \\ \cline{2-8} 
                                    & CPCL                              & 75.30                            & 86.63                            & 77.23                             & 57.11                            & 84.23                            & 94.12                                \\ \cline{2-8} 
                                    & DUEB                              & \textbf{75.94} & \textbf{87.08}  & \textbf{78.25}  & \textbf{57.46}  & \textbf{85.61} & \textbf{94.17}      \\ \hline
\multirow{3}{*}{1/4}                & Supervised                        & 71.66                           & 78.98                            & 74.31                             & 55.91                            & 82.05                            & 93.23                                \\ \cline{2-8} 
                                    & CPCL                              & 74.58                           & 85.39                            & 76.62                             & 56.69                            & 83.65                            & 93.70                                \\ \cline{2-8} 
                                    & DUEB                              & \textbf{75.85} & \textbf{86.69}  & \textbf{78.6}    & \textbf{57.13}  & \textbf{85.48}  & \textbf{94.22}      \\ \hline
\multirow{3}{*}{1/8}                & Supervised                        & 67.16                           & 73.47                            & 71.94                             & 48.66                            & 81.98                            & 92.28                                \\ \cline{2-8} 
                                    & CPCL                              & 73.74                           & 84.24                            & 76.53                             & 54.91                            & 84.36                            & 93.60                                 \\ \cline{2-8} 
                                    & DUEB                              & \textbf{74.89} & \textbf{85.47} & \textbf{78.75}   & \textbf{54.91} & \textbf{85.2}   & \textbf{94.14}      \\ \hline
\multirow{3}{*}{1/16}               & Supervised                        & 62.00                           & 67.65                            & 67.88                             & 41.56                            & 79.99                            & 91.63                                \\ \cline{2-8} 
                                    & CPCL                              & 71.66                           & 82.73                            & 76.24                             & \textbf{49.70}  & 83.57                            & 93.02                                \\ \cline{2-8} 
                                    & DUEB                              & \textbf{72.41} & \textbf{84.66}  & \textbf{77.09}   & {49.60}                        & \textbf{84.09}  & \textbf{93.32}      \\ \hline
\end{tabular}
}
\caption{Quantitative Performance of PASCAL VOC 2012 in percent over the Supervised and CPCL baselines with ResNet-50 backbone.(ANIMAL: Bird, Cat, Cow, Dog, Horse, Sheep; VEHICLE: Aeroplane, Bicycle, Boat, Bus, Car, Motorbike and Train; INDOOR: Bottle, Chair, Dining Table, Potted Plant, Sofa, and monitor; PERSON:Person; BACKGROUND: Background.}

\label{tab1}
\end{table*}
\begin{table*}[t]
 \footnotesize
\centering
\begin{tabular}{|cc|cc|cc|cc|}
\hline
\multicolumn{2}{|c|}{Partition} & \multicolumn{2}{c|}{1/4} & \multicolumn{2}{c|}{1/8} & \multicolumn{2}{c|}{1/16} \\ \hline
\multicolumn{2}{|c|}{Loss} & \multicolumn{2}{c|}{Datasets} & \multicolumn{2}{c|}{Datasets} & \multicolumn{2}{c|}{Datasets} \\ \hline
\multicolumn{1}{|c|}{Uncertainty} & Energy & \multicolumn{1}{c|}{PASCALVOC} & Cityscapes & \multicolumn{1}{c|}{PASCALVOC} & Cityscapes & \multicolumn{1}{c|}{PASCALVOC} & Cityscapes \\ \hline
\multicolumn{1}{|c|}{} &  & \multicolumn{1}{c|}{74.58} & 76.98 & \multicolumn{1}{c|}{73.74} & 74.60 & \multicolumn{1}{c|}{71.66} & 69.92 \\ \hline
\multicolumn{1}{|c|}{$\surd$} &  & \multicolumn{1}{c|}{75.23} & 77.28 & \multicolumn{1}{c|}{74.10} & 75.98 & \multicolumn{1}{c|}{71.90} & 72.07 \\ \hline
\multicolumn{1}{|c|}{} & $\surd$ & \multicolumn{1}{c|}{73.85} & 77.19 & \multicolumn{1}{c|}{73.00} & 75.80 & \multicolumn{1}{c|}{71.14} & 72.19 \\ \hline
\multicolumn{1}{|c|}{$\surd$} & $\surd$& \multicolumn{1}{c|}{\textbf{75.85}} & \textbf{77.85 }& \multicolumn{1}{c|}{\textbf{74.89}} & \textbf{76.16} & \multicolumn{1}{c|}{\textbf{72.41}} & \textbf{72.38}\\ \hline
\end{tabular}
\caption{mIoU of different loss components in addition to cross-entropy loss at different partitions.}
\label{tab:ab}
\vspace{-1em}
\end{table*}
 \subsection {Energy Modeling}
The objective of energy-based models is to incorporate generative modeling along with discriminative modeling. The energy-based model is also very well suited for uncertainty calibration for SS learning~\cite{zhao2020joint}. The generative model learns the joint distribution of the data and label $p(x,y)$, whereas the discriminative model learns conditional distribution $p(y|x)$. The log-likelihood distribution of joint distribution can be obtained from the discriminative function and energy of data distribution~\cite{jem}.
\begin{equation}
    log\ p_{\theta}(x,y)=log\ p_{\theta}(x) + log\ p_{\theta}(y|x)
\label{joint}
\end{equation}
In the above equation, conditional class distribution $log\ p_{\theta}(y|x)$ is learned by cross-entropy loss, and $log\ p_{\theta}(x)$ can be learned from the energy-based models.

Energy-based models originate from the fact that any probability distribution $p_{\theta}(x)$ can be expressed in terms of energy function with the Boltzmann distribution given by Eq.~\ref{energy}:
\begin{equation}
   p_{\theta}(x)= \frac{exp(-E_{\theta}(x))}{Z(\theta)}
\label{energy}   
\end{equation}
where $E_{\theta(x)}:\mathbb{R}^D\rightarrow\mathbb{R}$ is the energy function which maps $D$ dimension input into a scalar, and $Z_{\theta}=\int_{x}exp(-E_{\theta}(x))$ is the normalizing constant (partition function). The energy function value is low for the samples drawn from the data distribution and high otherwise ~\cite{towards}. The energy function can be defined in terms of the LogSumExp(.) of the logits of a classifier given by Eq.~\ref{ener}:
\begin{equation}
    E_{\theta}(x)= -LogSumExp_{y}(f_{\theta}(x)|y)
\label{ener}
\end{equation}
where $f_{\theta}(x)|y$ denotes the logits corresponding to $y$ class-label.

The energy-based loss $ \mathcal{L}_{e}$, which tends to maximize the logits over model distribution, is given by this equation:
\begin{equation}
    \mathcal{L}_{e}=LogSumExp_{y}(f_{\theta}(x)|y)
\label{eq:ene}
\end{equation}
From Eq.~\ref{eq:ene}, we can obtain the energy-based loss function for the conservative and progressive branch as $\mathcal{L}^c_{e}$ and $\mathcal{L}^p_{e}$ respectively. In our network, the energy-based loss is applied over conservative and progressive branches using intersection and union pseudo labels. It is also applied as a part of supervised loss using the ground-truth label.

\begin{figure*}[t]
        \centering
        \includegraphics[width=0.920\textwidth]{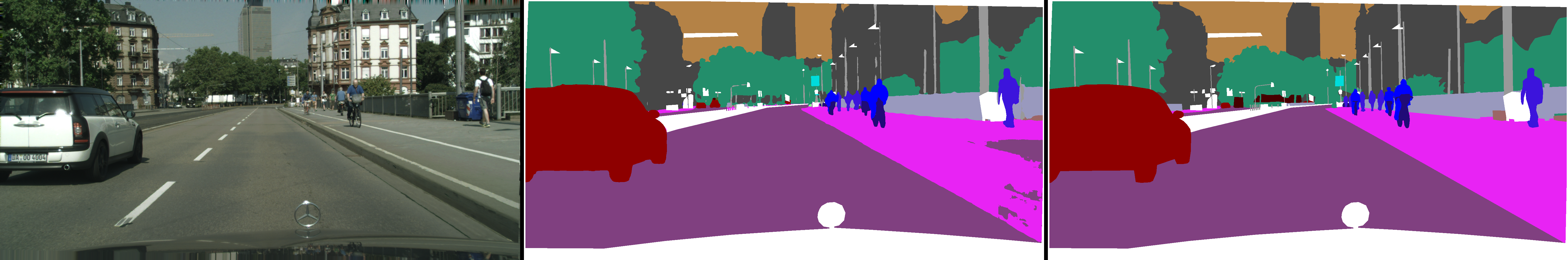}
        \includegraphics[width=0.920\textwidth]{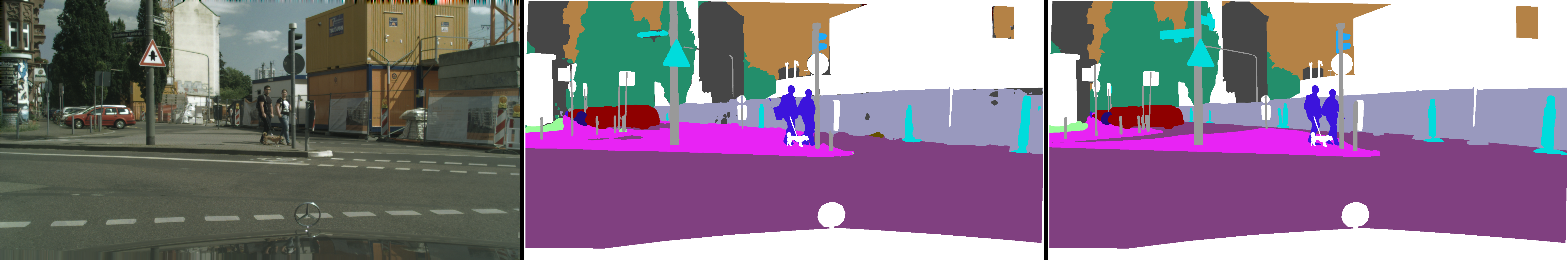}
        \includegraphics[width=0.920\textwidth]{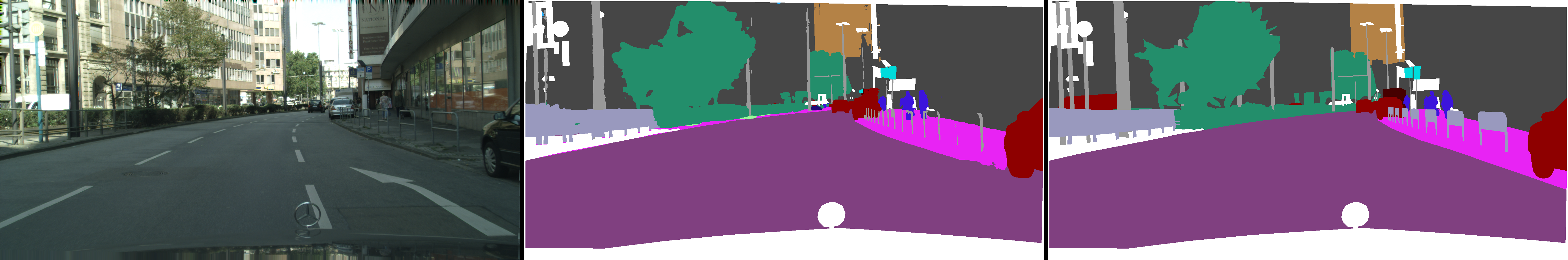}
        \caption{Segmentation Results of Cityscapes dataset (partition protocol:1/8) Left:Input Middle:Predictions of DUEB Right:Ground-truth.}
        \label{fig:city}
    \end{figure*}

\begin{table*}[]
\footnotesize
\centering
\begin{tabular}{|c|c|c|c|c|c|c|c|c|c|}
\hline
\textbf{Partition}         & \textbf{Methods} & \textbf{mIoU}  & \textbf{Flat}  & \textbf{Human} & \textbf{Vehicle} & \textbf{Construction} & \textbf{Object} & \textbf{Nature} & \textbf{Sky}   \\ \hline
\multirow{3}{*}{1/2}  & Supervised  & 75.36                           & 89.87                           & 71.71                           & 74.21                             & 67.35                                  & 69.91                            & 78.52                            & 94.52                           \\ \cline{2-10} 
                      & CPCL        & \textbf{78.17} & 91.44                           & \textbf{73.88} & \textbf{79.29}   & 69.15                                  & \textbf{72.44}  & \textbf{79.68}  & \textbf{94.72} \\ \cline{2-10} 
                      & DUEB        & 77.58                           & \textbf{91.58} & 71.97                           & 79.07                             & \textbf{69.23}        & 70.55                            & 79.24                            & 94.70                           \\ \hline
\multirow{3}{*}{1/4}  & Supervised  & 73.22                           & 88.88                           & 71.53                           & 70.4                              & 64.07                                  & 69.8                             & 75.98                            & 94.34                           \\ \cline{2-10} 
                      & CPCL        & 76.98                           & 90.85                           & 72.77                           & 77.4                              & 68.24                                  & 71.89                            & 78.06                            & 94.50                           \\ \cline{2-10} 
                      & DUEB        & \textbf{77.85} & \textbf{91.60} & \textbf{73.45} & \textbf{78.63}   & \textbf{70.16}        & \textbf{71.99} & \textbf{78.12}  & \textbf{94.72} \\ \hline
\multirow{3}{*}{1/8}  & Supervised  & 68.63                           & 87.89                           & 69.81                           & 60.02                             & 60.68                                  & 67.85                            & 74.43                            & 94.01                           \\ \cline{2-10} 
                      & CPCL        & 74.6                            & 89.48                           & 72.88                           & 72.89                             & 65.05                                  & 70.49                            & 77.13                            & 94.4                            \\ \cline{2-10} 
                      & DUEB        & \textbf{76.16} & \textbf{90.61} & \textbf{72.91} & \textbf{75.92}   & \textbf{65.53}        & \textbf{72.11}  & \textbf{78.42}  & \textbf{94.65} \\ \hline
\multirow{3}{*}{1/16} & Supervised  & 61.67                           & 85.27                           & 65.66                           & 46.88                             & 53.49                                  & 63.03                            & 72.71                            & 93.51                           \\ \cline{2-10} 
                      & CPCL        & 69.92                           & 88.41                           & 70.15                           & 61.96                             & 62.69                                  & 68.5                             & 75.97                            & 94.02                           \\ \cline{2-10} 
                      & DUEB        & \textbf{72.38} & \textbf{89.76} & \textbf{71.67} & \textbf{66.65}   & \textbf{64.38}        & \textbf{70.34}  & \textbf{77.10}   & \textbf{94.24} \\ \hline
\end{tabular}
\caption{Quantitative Performance of CityScapes dataset in percent over Supervised and CPCL baseline with ResNet-50 backbone (FLAT:Road and Sidewalk; HUMAN:Person and Rider; VEHICLE: Car, Truck, Bus, Train, MotorCycle, and Bicycle; CONSTRUCTION: Building Wall and Fence; OBJECT: Pole, Traffic Lights, and Traffic Sign, NATURE: Vegetation and Terrain; SKY: Sky )}
\label{tab:my-table}
\end{table*}

\subsection {Total loss}
The total loss is the combination of loss using the union and intersection label as pesdueo label defined in Eq.
\begin{equation}
    \mathcal{L}_{total}=\mathcal{L}_{det}+\gamma_{ale}(\mathcal{L}^c_{ale}+\mathcal{L}^p_{ale})+\gamma_e(\mathcal{L}^c_{e}+\mathcal{L}^p_{e})
\end{equation}
$\gamma_{ale}$ and $ \gamma_{e}$ are the hyperparameter for data uncertainty  and energy based loss functions. In our case we use both the value as 1.
\section{Experiments}
\label{sec:experiment}

\begin{table}[ht]
\centering
\scalebox{0.85}{
\begin{tabular}{|p{6.5 em}|c|c|c|c|c|}
\hline
\textbf{Methods} & \textbf{Network} & \textbf{1/16} & \textbf{1/8} & \textbf{1/4} & \textbf{1/2} \\ \hline
Supervised & ResNet-50 & 62.00 & 67.16 & 71.66 & 74.05 \\ \hline
\multirow{2}{*}{MT~\cite{mt}\textsubscript{\textit{NeurIPS'17}}} & ResNet-50 & 66.77 & 70.78 & 73.22 & 75.41 \\ \cline{2-6} 
 & ResNet-101 & 70.59 & 73.20 & 76.62 & 77.61 \\ \hline
\multirow{2}{*}{GCT\cite{gct}\textsubscript{\textit{ECCV'20}}} & ResNet-50 & 64.05 & 70.47 & 73.45 & 75.20 \\ \cline{2-6} 
 & ResNet-101 & 69.77 & 73.30 & 75.25 & 77.14 \\ \hline
\multirow{2}{*}{CCT~\cite{cct}\textsubscript{\textit{CVPR'20}}} & ResNet-50 & 65.22 & 70.87 & 73.43 & 74.75 \\ \cline{2-6} 
 & ResNet-101 & 67.94 & 73.00 & 76.17 & 77.56 \\ \hline
\multirow{2}{*}{\begin{tabular}[c]{@{}c@{}}Cut-Mix~\cite{cutmixseg}  \\ \textsubscript{\textit{BMVC'20}}\end{tabular}} & ResNet-50 & 68.9 & 70.70 & 72.46 & 74.49 \\ \cline{2-6} 
 & ResNet-101 & 72.56 & 72.69 & 74.25 & 75.89 \\ \hline

CPS~\cite{crosspseudo}\textsubscript{\textit{CVPR'21}}  & ResNet-50 & 68.21 & 73.20 & 74.24 & 75.91 \\ \hline
% PGC~\cite{kong2023pruning}\textsubscript{\textit{WACV'23}} & ResNet-50 & $-$ & \textbf{75.20} & \textbf{76.00} & $-$ \\ \hline
\multirow{2}{*}{CPCL ~\cite{fan2023}\textsubscript{\textit{TIP'23}}} & ResNet-50 & 71.66 & 73.74 & 74.58 & 75.30 \\ \cline{2-6} 
 & ResNet-101 & 73.44 & 76.40 & 77.16 & 77.67 \\ \hline

\multirow{2}{*}{Unimatch ~\cite{yang2023revisiting}} & ResNet-50 & 74.5 & 75.8 & 76.1 & $-$\\ \cline{2-6} 
 & ResNet-101 & 76.5 & 77.0 & 77.2 & $-$\\ \hline
 
DUEB & ResNet-50 & 72.41 & 74.89 & 75.85 & 75.94 \\ \hline
 DUEB & ResNet-101 & {74.13}   & {76.70} & {78.00}  &   {78.13}\\  \hline
\end{tabular}
}
\caption{Quantitative Performance (mIoU in percentage) of PASCAL-VOC dataset with other methods.}
\label{com}
\end{table}

% \subsection {Statistical Significance Analysis}   
%Figure \ref{fig:mat} and \ref{fig:voc} show the statistical significance ~\cite{stat} of the proposed method (DUEB) against supervised and CPCL baseline. The Nemenyi test is a post hoc test that is often used following a significant result in an analysis of variance (ANOVA) or a Friedman test, to determine which groups or treatments differ significantly from each other. The critical difference(CD) of the test is dependent on the confidence level (set to 0.05 for this exp) and average ranks for a number of datasets. If the difference between the the rank of two methods lies beyond the CD, then the methods are significantly different. It has been observed that DUEB is significantly different from CPCL and Supervised.

\begin{table}[ht]
\centering
\scalebox{0.82}{
\begin{tabular}{|p{6.5 em}|c|c|c|c|c|}
\hline
\textbf{Methods} & \textbf{Network} & \textbf{1/16} & \textbf{1/8} & \textbf{1/4} & \textbf{1/2} \\ \hline
Supervised & ResNet-50 & 61.67 & 68.63 & 73.22 & 75.36 \\ \hline
\multirow{2}{*}{MT~\cite{mt}\textsubscript{\textit{NeurIPS'17}}} & ResNet-50 & 66.14 & 72.03 & 74.47 & 77.43 \\ \cline{2-6} 
 & ResNet-101 & 68.08 & 73.71 & 76.53 & 78.59 \\ \hline
\multirow{2}{*}{GCT\cite{gct}\textsubscript{\textit{ECCV'20}}} & ResNet-50 & 65.81 & 71.33 & 75.30 & 77.09 \\ \cline{2-6} 
 & ResNet-101 & 66.90 & 72.96 & 76.45 & 78.58 \\ \hline

\multirow{2}{*}{CCT~\cite{cct} \textsubscript{\textit{CVPR'20}}} & ResNet-50 & 66.35 & 72.46 & 75.68 & 76.78 \\ \cline{2-6} 
 & ResNet-101 & 69.64 & 74.48 & 76.35 & 78.29 \\ \hline
CPS~\cite{crosspseudo}\textsubscript{\textit{CVPR'21}} & ResNet-50 & 69.79 & 74.39 & 76.85 & 78.64 \\ \hline
PGC~\cite{kong2023pruning}\textsubscript{\textit{WACV'23}}  & ResNet-50 & $-$ & 71.20 & 73.90 & 76.80 \\ \hline
CPCL~\cite{fan2023}\textsubscript{\textit{TIP'23}} & ResNet-50 & 69.92 & 74.60 & 76.98 & 78.17 \\ \hline

\multirow{2}{*}{Unimatch~\cite{yang2023revisiting}} & ResNet-50 & 75.0 & 76.8 & 77.5 & 78.6 \\ \cline{2-6} 
 & ResNet-101 & 76.6 & 77.9 & 79.2 & 79.5 \\ \hline
 
DUEB & ResNet-50 & 72.38 & 76.16 & 77.85 & 77.58 \\ \hline
DUEB & ResNet-101 & {74.16} & {77.36} &{ 78.80} & {79.47} \\ \hline
\end{tabular}
}
\caption{Quantitative Performance (mIoU in percentage) 
of CityScapes dataset with other methods.}
\label{comp2}
\end{table}

We evaluate the performance of the proposed method on two standard benchmark datasets: Cityscapes~\cite{cityscapes} and PASCAL VOC 2012~\cite{PASCAL}. The experiments are performed using the PyTorch deep learning framework on a server with an NVIDIA A100. The detailed algorithm and training details are presented in the Supplementary materials. 
\subsection{  Dataset and Implementation Details}
\subsubsection{Datasets}
\vspace{-0.5em}
The Cityscapes dataset is formed explicitly for the urban driving scene. It comprises 2975 training images and 500 validation images. The resolution of each image is $2048\times1024$, which is randomly cropped to $800\times800$, keeping the original resolution intact. There are 19 classes for pixel annotation.

PASCAL VOC 2012 dataset is the benchmark semantic segmentation dataset of the common objects. It has 1464 training images and 1449 validation images, with twenty foreground classes and one background class. We have created the augmented version of the dataset using the Segmentation Boundaries Dataset following the convention of the previous work~\cite{fan2023}. Therefore, the entire training set now contains 10582 images with random cropping of $512\times512$, keeping the original resolution intact.
\vspace{-1em}
\subsubsection{Implementation Details}
We follow a similar experimental protocol defined in the CPCL model for a fair comparison. Both the conservative and progressive branch share the same structure (Deeplabv3+) with randomly initialized segmentation heads. We use SGD with momentum with an initial learning rate of $10^{-4}$ and $5 \times 10^{-3}$ with Cityscapes and PASCAL VOC, respectively. The learning rate is multiplied by $(1-\frac{iter}{max_iter})^{0.9}$, momentum is set to $0.9$, and the weight decay is $10^{-4}$. The data partition protocol divides the data into labeled and unlabeled sets. It randomly extracts 1/2, 1/4, 1/8, and 1/16 of the whole data labeled part and considers the remaining data as unlabeled part. We follow the data partition protocol of previous works, \cite{fan2023,gct}. The Cut-Mix strategy is used for the generation of strongly augmented images using three rectangle regions of random ratio (in the range of [0.25, 0.5]). The Monte Carlo sample value $T$ is set to 10 for both datasets. The implementation code will be made public.
\subsection{Performance Analysis}
We evaluate our framework DUEB on PASCAL VOC and Cityscapes dataset at all the partition protocols (ratio of labeled data). It has been observed that there is a significant improvement in mIoU at all partition protocols for the PASCALVOC dataset, as given in Table~\ref{tab1}. The performance analysis is done with CPCL and supervised CPCL baseline~\cite{fan2023}. It has been observed that there is a significant improvement by DUEB over CPCL supervised baseline as the ratio of labeled data decreases. The performance gains with respect to the supervised CPCL baseline is \textbf{1.89\%}, \textbf{4.19\%}, \textbf{7.73\%}, and \textbf{10.41\%} under partition protocol of 1/2, 1/4, 1/8 and 1/16, respectively on PASCAL VOC using ResNet-50 backbone. The best performance is over the animal class of PASCAL VOC, as there is an increment of about 17 \% over the supervised baseline and ~2\% over the CPCL network with ResNet-50 backbone. However, the performance is stagnant when it comes to background class as higher mIoU is already achieved by supervised baseline.

The CityScapes dataset also has significant mIoU improvement over partition protocols 1/4, 1/8 and 1/16, as given in Table~\ref{tab:my-table}. However, at 1/2 data partition, it outperformed the supervised CPCL baseline but showed a reduction slight reduction with respect to CPCL with ResNet-50. In Cityscapes, too, there is an improvement in performance as the ratio of labeled images decreases. The best performance is achieved by the vehicle group, which is 19.77\% over the supervised CPCL baseline and 4.69\% over CPCL. However, the performance is stagnant when it comes to sky class as higher mIoU is already achieved by supervised baseline. The visualization of the segmentation map is provided in Figure~\ref{fig:city}. It has been observed that DUEB with ResNet-101 backbone outperforms the state-of-the-art methods for all partition protocols as given in Table~\ref{tab1} and Table \ref{tab:my-table}. More details are available in the Supplementary material.

\subsection {Ablation Study}
We analyze the network to evaluate mIoU on separate loss components. In the entire network, standard cross entropy is loss with ground truth for the supervised part and with pseudo-intersection-union labels for the unsupervised part. Data uncertainty loss and energy-based loss have been added in addition to cross-entropy loss. From  Table~\ref{tab:ab} it has been observed that data uncertainty loss is more robust than the energy-based loss when used individually. However, a combination of both uncertainty-based loss and energy-based loss performs better in terms of mIoU than the individual components. The reason is energy-based modeling helps to quantify the aleatoric uncertainty because it is robust to noisy data and aims to minimize energy for plausible data points.

\subsection{Comparative Analysis}\
Table~\ref{com} and Table~\ref{comp2} gives a comparative analysis of DUEB on PASCAL-VOC and CityScapes with state-of-the art methods: GCT~\cite{gct}, CCT~\cite{cct}, Cut-Mix Seg,~\cite{cutmixseg}, MT~\cite{mt}, CPS~\cite{crosspseudo}, PGC~\cite{kong2023pruning}, CPCL~\cite{fan2023}, ~Unimatch~\cite{yang2023revisiting}. All compared methods are based on DeepLabv3+ with ResNet-50 or ResNet-101~\cite{he2016deep} backbone. We report the results of some previous methods in Table~\ref{com} and Table~\ref{comp2} from~\cite{crosspseudo} and \cite{fan2023}. It has outperformed some previous methods on both ResNet-50/101 backbone in terms of mIoU. The performance improvement is more evident when a few number of labeled images are used.
%\vspace{-1em}
% \begin{table*}[t]
%  \footnotesize
% \centering
% \begin{tabular}{|cc|cc|cc|cc|}
% \hline
% \multicolumn{2}{|c|}{Partition} & \multicolumn{2}{c|}{1/4} & \multicolumn{2}{c|}{1/8} & \multicolumn{2}{c|}{1/16} \\ \hline
% \multicolumn{2}{|c|}{Loss} & \multicolumn{2}{c|}{Datasets} & \multicolumn{2}{c|}{Datasets} & \multicolumn{2}{c|}{Datasets} \\ \hline
% \multicolumn{1}{|c|}{Uncertainty} & Energy & \multicolumn{1}{c|}{PASCALVOC} & Cityscapes & \multicolumn{1}{c|}{PASCALVOC} & Cityscapes & \multicolumn{1}{c|}{PASCALVOC} & Cityscapes \\ \hline
% \multicolumn{1}{|c|}{} &  & \multicolumn{1}{c|}{74.58} & 76.98 & \multicolumn{1}{c|}{73.74} & 74.60 & \multicolumn{1}{c|}{71.66} & 69.92 \\ \hline
% \multicolumn{1}{|c|}{$\surd$} &  & \multicolumn{1}{c|}{75.23} & 77.28 & \multicolumn{1}{c|}{74.10} & 75.98 & \multicolumn{1}{c|}{71.90} & 72.07 \\ \hline
% \multicolumn{1}{|c|}{} & $\surd$ & \multicolumn{1}{c|}{73.85} & 77.19 & \multicolumn{1}{c|}{73.00} & 75.80 & \multicolumn{1}{c|}{71.14} & 72.19 \\ \hline
% \multicolumn{1}{|c|}{$\surd$} & $\surd$& \multicolumn{1}{c|}{\textbf{75.85}} & \textbf{77.85 }& \multicolumn{1}{c|}{\textbf{74.89}} & \textbf{76.16} & \multicolumn{1}{c|}{\textbf{72.41}} & \textbf{72.38}\\ \hline
% \end{tabular}
% \caption{mIoU of different loss components in addition to cross-entropy loss at different partitions.}
% \label{tab:ab}
% \vspace{-1em}
% \end{table*}
\section{Conclusion}
\label{sec:conclusion}
We propose data uncertainty and energy-based modeling for SS semantic segmentation. The per-pixel variance parameter, along with the energy, have resulted in performance improvement. Moreover, the usage of union-intersection pseudo labels for unsupervised parts renders usage of both large quantity and high quality. The proposed methodology is generic, i.e., it can be applied to existing SS semantic segmentation networks. The modification in the architecture of CNN to incorporate the output variance layer can yield improved results. However, there is further scope for improvement in the areas of pseudo-label generation, uncertainty loss incorporation in transformers, and model coupling. \\
\textbf{Acknowledgments} Rini acknowledges the financial support provided by the Department of Science and Technology, Government of India, through the WISE Post-Doctoral Fellowship Program (Reference No. DST/WISE-PDF/ET-33/2023), which facilitated the successful completion of this work.
%%%%%%%%% REFERENCES
{\small
\bibliographystyle{ieee_fullname}
\bibliography{egbib}
}

\end{document}

%%%%%%%%% TITLE
% \title{\ Data Uncertainty and Energy based Semi-Supervised Semantic Segmentation}

\title{Supplementary: Uncertainty and Energy based Loss Guided Semi-Supervised Semantic Segmentation}

\author{Rini Smita Thakur, Vinod K Kurmi\\
Indian Institute of Science Education and Research, Bhopal, India\\
%Institution1 address\\
{\tt\small \{rinithakur, vinodkk\}@iiserb.ac.in}
% For a paper whose authors are all at the same institution,
% omit the following lines up until the closing ``}''.
% Additional authors and addresses can be added with ``\and'',
% just like the second author.
% To save space, use either the email address or home page, not both
% \and
% Vinod K Kurmi\\
% IISER, Bhopal\\
% %First line of institution2 address\\
% {\tt\small vinodkk@iiserb.ac.in}
}

\maketitle
\thispagestyle{empty}

%%%%%%%%% ABSTRACT
% \begin{abstract}
 
% \end{abstract}

%%%%%%%%% BODY TEXT
In this supplementary, we present the experimental details using algorithm flow. We also provide additional results on  PASCAL-VOC dataset  and compare with other methods on ResNet101 on 1/8 partition. We provide a segmentation visualization and comparison with baseline methods.

\section{Algorithm}
In this section, we provide a comprehensive explanation of the algorithms employed to train our segmentation model using uncertainty and energy-based loss. While the main paper discusses the details of the loss function, this entire algorithmic process is given in Algorithm 1.

\begin{figure}[h]
        \centering
        \includegraphics[width=0.45\textwidth]{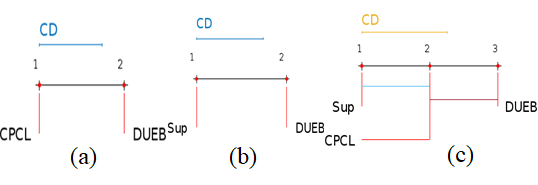}
        \vspace{-0.5em}
        \caption{Analysis of statistical significant difference for Supervised, CPCL and DUEB model on CityScapes Dataset at 1/8 partition (a) CPCL and DUEB (b) Sup and DUEB (c) Sup, CPCL and DUEB}
        \label{fig:mat}
        \vspace{-1.5em}
    \end{figure}

 \begin{figure}[h]
        \centering
        \includegraphics[width=0.45\textwidth]{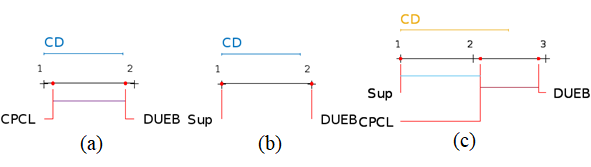}
        \vspace{-0.5em}
        \caption{Analysis of statistical significant difference for Supervised, CPCL and DUEB model on PASCAL VOC Dataset at 1/8 partition (a) CPCL and DUEB (b) Sup and DUEB (c) Sup, CPCL and DUEB}
        \label{fig:voc}
        \vspace{-1em}
    \end{figure}
\begin{algorithm*}[!]
\SetAlgoLined
\KwIn{
    Training dataset labeled data $D_{l}=\left\{(X_{l}^{1}, 
G^{1}),.......,(X_{l}^{N}, G^{N})\right\}$  and 
unlabeled data $D_{u}=\left\{X_{u}^{1}, .......,X_{u}^{Q}\right\}$
}
\KwOut{Trained segmentation model $f(;\theta)$}
\For{mini batch of labeled samples $X_l^i, G^i\in D_{l}$ or $X^j, X^k\in D_{u}$  }{
$X^{jk}_u \leftarrow$ mix($X^j,X^k,mask$)\;
\textbf{Supervised loss:} \\
$y_{lc}^i$ $\leftarrow$ $f(X_l^i;\theta_c)$ ; 
$ \mathcal{L}_{sup} $ $\leftarrow$   $\operatorname{CE} (y_{lc}^i;G_l^i) $

\textbf{Unsupervised loss: }\\
For input $X^{jk}_i $ obtained the $y_{\text {inter }}^i$ , $ y_{\text {union}}^i$ and $w_{\text {u}}^i$ using \cite{fan2023} \\

$y_{uc}^i, \sigma_{uc}^i$ $\leftarrow$ $f(X^{jk}_u;\theta_c)$ ; and 
$y_{up}^i ,\sigma_{up}^i$ $\leftarrow$ $f(X^{jk}_u;\theta_p)$ \\

$ \mathcal{L}_{int} $ $\leftarrow$   $w_{\text {u}}^i$$\operatorname{CE} (y_{uc}^i;y^i_{\text {inter}}) $ ; and 
$ \mathcal{L}_{uni} $ $\leftarrow$   $w_{\text {u}}^i$$\operatorname{CE} (y_{up}^i;y^i_{\text {union}}) $ \\
$ \mathcal{L}_{det}\leftarrow \mathcal{L}_{sup}+\gamma_{int}\mathcal{L}_{int}+\gamma_{uni}\mathcal{L}_{uni}$ \\
\textbf{ Uncertainty estimation: }\\

$\operatorname{\mathit{diff}_p} =\operatorname{CE} (y_{up}^i;y^i_{\text {union}})-\operatorname{CE} (y_{up}^i;y^i_{\text {union}}) \text {; }   \epsilon^p_{\mathrm{t}} \sim \mathcal{N}(0, \sigma_p)$ \\
$\operatorname{\mathit{diff}_c} =\operatorname{CE} (y_{uc}^i;y^i_{\text {inter}})-\operatorname{CE} (y_{up}^i;y^i_{\text {inter}}) \text {; }   \epsilon^c_{\mathrm{t}} \sim \mathcal{N}(0, \sigma_c)$ \\
Using Eq.(12) from main paper, obtained $\operatorname{\mathcal{L}_{ale}^c}$ and $\operatorname{\mathcal{L}_{ale}^p}$
\\
\textbf{Energy loss:} \\
$ \mathcal{L}_{e}^c $ $\leftarrow$ $ LogSumExp_{y^i_{\text {inter}}}(f(X^{jk}_u;\theta_c)|y^i_{\text {inter}})$ \\
$ \mathcal{L}_{e}^p $ $\leftarrow$ $ LogSumExp_{y^i_{\text {union}}}(f(X^{jk}_u;\theta_p)|y^i_{\text {union}})$ \\
Total loss: 
$\mathcal{L}_{total}$ $\leftarrow$ $ \mathcal{L}_{det}+\gamma_{ale}(\mathcal{L}^c_{ale}+\mathcal{L}^p_{ale})+\gamma_e(\mathcal{L}^c_{e}+\mathcal{L}^p_{e})$ \\
Update parameters: \\
$\hat{\theta_c}$ $\leftarrow$ $\theta_c$ -$\lambda\frac{\partial\mathcal{L}_{total}}{\partial \theta_c}$ ; and 
$\hat{\theta_p}$ $\leftarrow$ $\theta_p$ -$\lambda\frac{\partial\mathcal{L}_{total}}{\partial \theta_p}$ \\ 
       }

Output trained segmentation parameters $\hat{\theta_c}$
\caption{Semi-Supervised Semantic Segmentation using  Uncertainty and Energy-Based Loss in Pseudo Labels}
\label{one}
\end{algorithm*}

\begin{figure*}
    \centering

    \begin{subfigure}[b]{0.23\textwidth}
        \includegraphics[width=\textwidth]{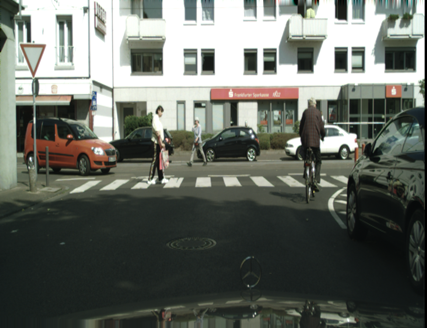}
        \caption{Input}
    \end{subfigure}
    \hfill
    \begin{subfigure}[b]{0.23\textwidth}
        \includegraphics[width=\textwidth]{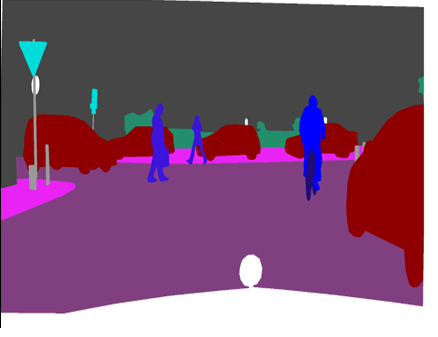}
        \caption{Ground-truth}
    \end{subfigure}
    \hfill
    \begin{subfigure}[b]{0.23\textwidth}
        \includegraphics[width=\textwidth]{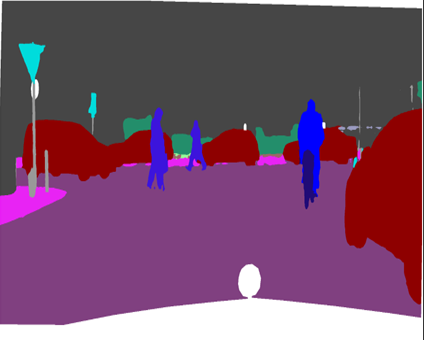}
        \caption{CPCL}
    \end{subfigure}
    \hfill
    \begin{subfigure}[b]{0.23\textwidth}
        \includegraphics[width=\textwidth]{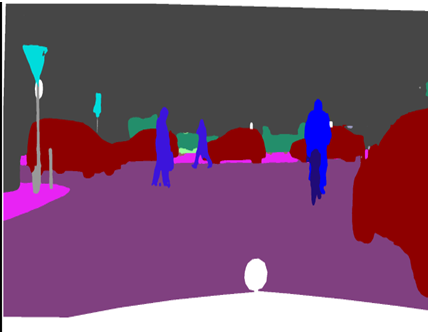}
        \caption{DUEB}
         
    \end{subfigure}
    \caption{Segmentation Results of Cityscapes dataset (partition protocol: 1/8). The missegmented pixels in right part of image (gray) in CPCL are rectified on DUEB.}
    \label{fig:comparative}
\end{figure*}

\section{Visualization}
Furthermore, we provide visualizations of the segmentation outputs and compare them with baseline methods. These visualizations allow for a qualitative assessment of the segmentation quality achieved by our proposed method.
Figure~\ref{fig:comparative} displays the segmentation results obtained from the Cityscapes dataset, using a partition protocol of 1/8. We compare our results with the state-of-the-art method CPCL. Upon observing the figure, it becomes evident that the CPCL method exhibits false positives for the "Pole" class. Specifically, in the bottom-right image, the right portion mistakenly identifies a pole, whereas the ground truth does not include a pole at that location. In contrast, our proposed method's predictions align more accurately with the ground truth.

\section {Statistical Significance Analysis}   
Figure \ref{fig:mat} and \ref{fig:voc} show the statistical significance ~\cite{stat} of the proposed method (DUEB) against supervised and CPCL baseline. The Nemenyi test is a post hoc test that is often used following a significant result in an analysis of variance (ANOVA) or a Friedman test, to determine which groups or treatments differ significantly from each other. The critical difference(CD) of the test is dependent on the confidence level (set to 0.05 for this exp) and average ranks for a number of datasets. If the difference between the the rank of two methods lies beyond the CD, then the methods are significantly different. It has been observed that DUEB is significantly different from CPCL and Supervised.

\section {Class Disagreement Indicator}
The detailed explanation of class disagreement indicator (I) utilized for union pseudolabels is given in this section. When there is disagreement between class prediction of two branches, the conventional approach selects pixel with high confidence as pseudolabel. However, this approach can be erroneous as the prediction with higher confidence can be wrong. Therefore, disagreement indicator selects the difficult class which is likely to create more confusion. Class disagreement indicator obtained by agreement matrix ($M \in \mathbb{R}^{C \times C}$)
 is used for class selection. The agreement matrix is $C \times C$ matrix with the entries $m_{j,k}$, which denotes that number of pixels where $Y_{cw}$ is from class (j) and $Y_{pw}$ is from class (k). The steps of disagreement indicator calculation are:\\
 1. Calculation of agreement matrix ($M \in \mathbb{R}^{C \times C}$), where C is total number of classes. $m_{j,k}$ denotes number of pixels where $Y_{cw}$ is from class (j) and $Y_{pw}$ is from class (k) and $j, k \in [1, C]$. \\
 2. Calculation of class disagreement indicator 
\begin{align}
I_j &= 2 - \frac{m_{j,j}}{\sum_{k=1}^{C} m_{j,k}} - \frac{m_{j,j}}{\sum_{k=1}^{C} m_{k,j}} \\
I_k &= 2 - \frac{m_{k,k}}{\sum_{j=1}^{C} m_{k,j}} - \frac{m_{k,k}}{\sum_{j=1}^{C} m_{j,k}}
\end{align}
$\text{where } j, k \in [1, C] \text{ are the indices of the classes.}$ \\
3. Calculation of pseudolabel for disagreement part $l_d^i$:
\begin{align}
l_d^i &= c_j \quad \text{if } I_j \geq I_k, \, j \neq k \\
l_d^i &= c_k \quad \text{if } I_k \geq I_j, \, j \neq k
\end{align}
The union pseudolabel is combination of agreement and disagreement parts of two branches. The agreement part implies that both branches give same class prediction. The disagreement part pseudolabels are given with $l_d^i$. 
\section {Confidence based dynamic loss}
The presence of noise in pseudolabels is inevitable. The confidence based loss can further aid in network training with unreliable pseudolabels. It performs loss re-weighting based on confidence. The maximum softmax probability denotes class wise confidence. Let $b_c^i$ is the prediction confidence of the conservative branch at the $i^{th}$ pixel, similarly, $b_p^i$ is the prediction confidence of the conservative branch at the $i^{th}$ pixel. The confidence based loss is given by:
\begin{align}
\omega_u^i = \begin{cases}
\frac{1}{2}\left(b_c^i + b_p^i\right) & \text{if } Y_{cw}^i = Y_{pw}^i \\
b_c^i & \text{if } Y_{cw}^i \neq Y_{pw}^i, \, l_d^i \leftarrow Y_{cw}^i \\
b_p^i & \text{if } Y_{cw}^i \neq Y_{pw}^i, \, l_d^i \leftarrow Y_{pw}^i
\end{cases}
\end{align}
Therefore, confidence based loss reduces the impact of unreliable pseudolabel with low confidence.

% \begin{table*}[!b]
% \centering
% \begin{tabular}{|c|c|c|c|c|c|c|c|}
% \hline
% Methods    & Supervised & GCT~\cite{gct} & CCT~\cite{cct} & Cut-Mix Seg~\cite{cutmixseg} & MT ~\cite{mt} & CPCL~\cite{fan2023} & DUEB (ours) \\ \hline
% mIoU in \% &   68.38   &  73.3    & 73    &  72.69   & 73.2   &  76.40   &   \textbf{  76.70}   \\ \hline
% \end{tabular}
% \caption{Quantitative Performance (mIoU in percentage)
% of PASCAL-VOC dataset  with other methods on ResNet101 on 1/8 partition of the data}
% \label{tab:res}
% \end{table*}

{\small
\bibliographystyle{ieee_fullname}
\bibliography{egbib}
}